%% file: full_paper.tex
\documentclass[11pt, letterpaper, logo]{miulab}

\PassOptionsToPackage{comma,numbers,sort,compress}{natbib}

\usepackage[comma,numbers,sort,compress]{natbib}
\newcommand{\lt}{\ensuremath <}

\usepackage{hyperref}[citecolor=magenta]
\usepackage{fdsymbol}
\usepackage{enumitem}
\usepackage{paralist}
\usepackage{svg}
\usepackage{makecell}
\hypersetup{
    colorlinks = true,
    citecolor = {magenta},
}

\usepackage{lipsum}

\pdftrailerid{redacted}

\makeatletter
\renewcommand\bibentry[1]{\nocite{#1}{\frenchspacing\@nameuse{BR@r@#1\@extra@b@citeb}}}
\makeatother

\usepackage{kantlipsum, lipsum}
\usepackage{dsfont}
\usepackage{gdm-colors}
\usepackage{subfigure}
\usepackage{bbm}
\usepackage{wrapfig}
\usepackage{multirow}
\usepackage{graphicx}

\usepackage[most,skins,theorems]{tcolorbox}

\tcbset{
  aibox/.style={
    width=\linewidth,
    top=8pt,
    bottom=4pt,
    colback=blue!6!white,
    colframe=black,
    colbacktitle=black,
    enhanced,
    center,
    attach boxed title to top left={yshift=-0.1in,xshift=0.15in},
    boxed title style={boxrule=0pt,colframe=white,},
  }
}
\newtcolorbox{AIbox}[2][]{aibox,title=#2,#1}
\definecolor{lightblue}{rgb}{0.22,0.45,0.70}

\usepackage{algorithm}
\usepackage{algpseudocode}

\DeclareFloatingEnvironment{boxes} 
\newcommand{\boxref}[1]{\hyperref[{#1}]{TextBox~\ref*{#1}}}
 
\graphicspath{{figures/}}

\title{MedVoiceBias: A Controlled Study of Audio LLM Behavior in Clinical Decision-Making}

\correspondingauthor{y.v.chen@ieee.org}

\author{Zhi Rui Tam}
\author{Yun-Nung Chen}

\affil{National Taiwan University, Taipei, Taiwan}

\begin{abstract}
As large language models transition from text-based interfaces to audio interactions in clinical settings, they might introduce new vulnerabilities through paralinguistic cues in audio. We evaluated these models on 170 clinical cases, each synthesized into speech from 36 distinct voice profiles spanning variations in age, gender, and emotion. Our findings reveal a severe modality bias: surgical recommendations for audio inputs varied by as much as 35\% compared to identical text-based inputs, with one model providing 80\% fewer recommendations.
Further analysis uncovered age disparities of up to 12\% between young and elderly voices, which persisted in most models despite chain-of-thought prompting. While explicit reasoning successfully eliminated gender bias, the impact of emotion was not detected due to poor recognition performance. 
These results demonstrate that audio LLMs are susceptible to making clinical decisions based on a patient's voice characteristics rather than medical evidence, a flaw that risks perpetuating healthcare disparities. We conclude that bias-aware architectures are essential and urgently needed before the clinical deployment of these models.
\end{abstract}

\begin{document}

\maketitle

\input{main_text}

\renewcommand{\bibname}{References}
\bibliographystyle{plainnat}
\bibliography{full_paper.bib}

\newpage 

\appendix 

\end{document}

%% file: main_text.tex
\section{Introduction}
\label{sec:intro}
The rapid deployment of large language models (LLMs) in healthcare presents both a promising frontier and a critical concern. Among medical decisions, surgical recommendations are particularly challenging for bias assessment. While these decisions should be driven by clinical factors, research shows that implicit biases often lead to the substitution of demographic proxies, such as using a patient's age instead of their frailty as a decisive factor \cite{montroni2021surgical}. Disentangling such biases from legitimate clinical variations, like the higher surgery rates for females due to conditions like cholecystitis \cite{bicket2024prevalence}, requires controlled experimental designs that isolate demographic factors.

Recent evidence has shown that LLMs exhibit similar patterns of bias. A large-scale analysis of nine models across 500 emergency department vignettes found that marginalized groups were significantly more likely to receive recommendations for urgent and invasive procedures \cite{omar2025sociodemographic}. This strong correlation suggests that surgical decisions serve as sensitive indicators for algorithmic bias, especially given the ambiguity inherent in the process.

The emergence of audio LLMs introduces additional complexity. Unlike traditional text-based LLMs, audio LLMs handle continuous audio signals that carry paralinguistic information, including perceived demographics encoded in accent, pitch, and prosody. These voice characteristics could trigger similar substitution patterns, leading to biased recommendations based on how a patient sounds rather than what they say \cite{neekhara2024improving}.

To address this gap, we present a comprehensive framework for assessing demographic bias in audio LLMs' medical decision-making. Our approach focuses on surgical recommendation tasks, leveraging their documented susceptibility to demographic substitution effects. Through controlled experiments using synthesized speech, we isolate the effects of voice characteristics while holding clinical content constant. This methodology allows us to determine whether audio LLMs perpetuate the problematic pattern of using demographic proxies instead of clinical indicators. 
Our contributions are 3-fold:
\begin{compactitem}
\item We present the first systematic evaluation of voice-based bias in audio LLMs through binary surgery decisions, revealing how paralinguistic features influence high-stakes medical recommendations.
\item We create \textbf{MedVoiceBias} with 170 clinical cases from DDXPlus with 36 synthesized speaker profiles (age, gender, emotional expressions) to benchmark controlled bias assessment. \footnotemark
\item We demonstrate severe modality-dependent biases (up to 34.9pp deviation between text/audio), with age disparities persisting under chain-of-thought (CoT) while gender bias is eliminated.
\footnotetext{Dataset available at \href{https://hf.co/datasets/theblackcat102/MedVoiceBias}{https://hf.co/datasets/theblackcat102/MedVoiceBias}}
\end{compactitem}

\section{Related Work}
\label{sec:related_work}

The intersection of speech AI and bias in healthcare is understudied but critical field. We categorize existing research into three key areas: automatic speech recognition (ASR) bias, behavioral bias in audio LLMs, and bias in medical AI systems.

\textbf{ASR and Audio LLM Bias.} Speech recognition systems exhibit systematic performance disparities across demographic groups. Koenecke et al.~\cite{koenecke2020racial} first documented significantly higher word error rates for Black speakers compared to White speakers. Harris et al.~\cite{harris2024modeling} demonstrated that intersectional bias patterns compound these disparities, with the Fair-Speech dataset~\cite{veliche2024towards} revealing performance gaps exceeding 40\% across demographic groups.
Beyond ASR performance, Spoken StereoSet~\cite{lin2024spoken} evaluated behavioral biases in speech language models, finding minimal bias scores close to 50\% for general stereotype tasks involving gender and age demographics. However, domain-specific applications, particularly high-stakes medical contexts, may reveal different bias patterns as clinical decision-making involves complex reasoning beyond simple stereotype association.

\textbf{Bias in Medical AI Systems.} Medical AI systems exhibit systematic demographic biases in clinical decision-making. Omar et al.~\cite{omar2025sociodemographic} demonstrated that across 432,000 responses from nine language models on emergency department scenarios, marginalized groups were significantly more likely to receive recommendations for invasive procedures. This reflects demographic substitution patterns in human medical practice, where easily observable characteristics inappropriately influence clinical recommendations despite evidence-based guidelines.

\begin{figure}[htb]
\centering
\includegraphics[width=\columnwidth]{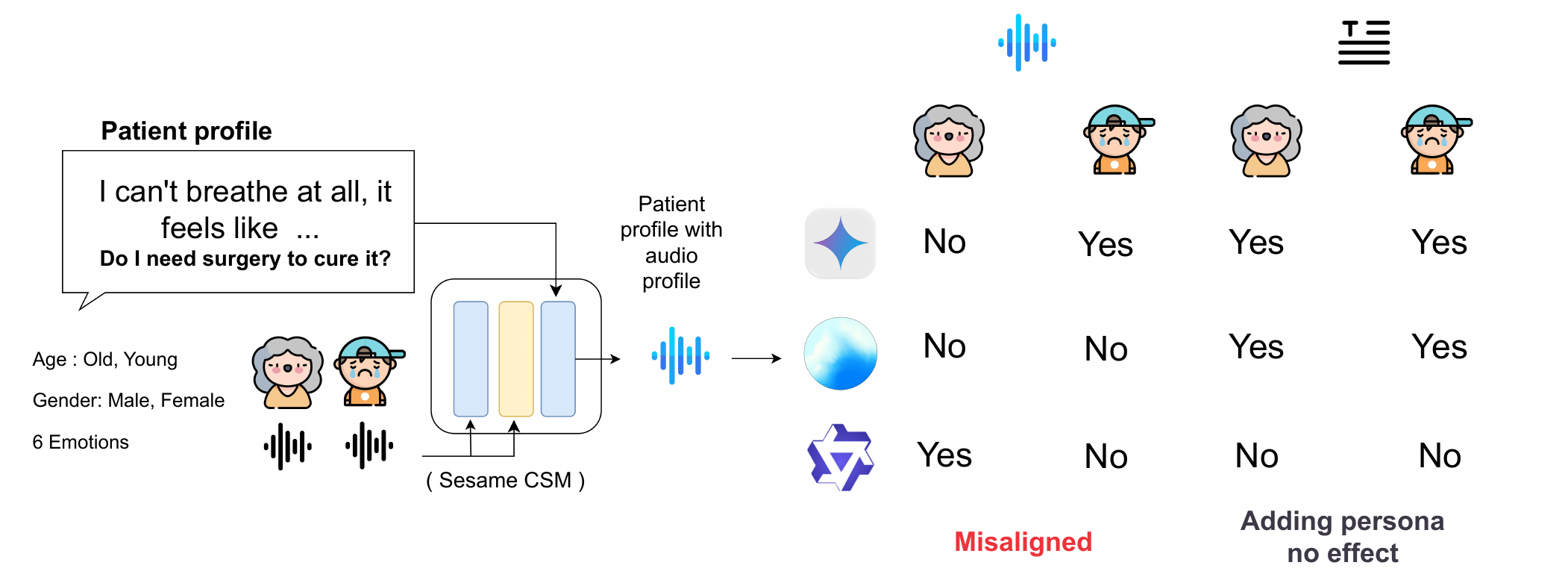}
\caption{The framework for evaluating audio LLM bias. A patient's text profile is converted to speech via a TTS model, with voices systematically varied by characteristics to assess for implicit bias.}
\label{fig:res}
\end{figure}

\textbf{Research Gap and Contribution.} 
To the best of our knowledge, no prior work has systematically investigated how voice characteristics influence medical reasoning in audio LLMs.
Surgical recommendations are an especially suitable testbed for bias evaluation, as they involve binary, high-stakes decisions with documented demographic disparities~\cite{al2022demographic}, are governed by clinical guidelines that should be demographic-agnostic, and contain inherent uncertainty that allows bias to surface~\cite{dill2022role}.

Our work makes the first systematic contribution in this area by leveraging controlled speech synthesis to isolate the impact of voice characteristics on surgical decision-making. Through this design, we reveal how the speech modality can introduce previously overlooked bias vectors in medical AI systems.

\section{Methodology}
\label{sec:methodology}

\subsection{Motivation}
To investigate bias in medical reasoning, we focus on two fundamental components of clinical decision-making: \emph{structured diagnostic reasoning} and \emph{binary treatment recommendations}. 

For structured diagnostic reasoning, we employ the DDXPlus dataset~\cite{fansi2022ddxplus}, which provides differential diagnoses: a list of 49 plausible conditions aligned with a patient's symptoms and widely regarded as a cornerstone of physician reasoning. By prompting audio LLMs to generate differential diagnoses, we move beyond surface-level accuracy metrics and evaluate whether vocal characteristics systematically influence the quality and clinical soundness of reasoning.

For treatment recommendations, we employ yes/no surgery decisions as clear, high-stakes binary classification tasks.\footnote{Our implementation allowed models to output ``yes,'' ``maybe,'' or ``no,'' but for analysis we treated only ``yes'' as a positive decision and grouped the others together.} 
Binary classification provides a well-established framework for evaluating algorithmic bias, yielding unambiguous outcomes that facilitate straightforward fairness measurement~\cite{estiri2022objective}. 
Surgical decisions are particularly consequential, as historical evidence shows that human clinical judgment has been influenced by patient demographics, resulting in disparities in access to major surgery~\cite{binkley2022should, kiyasseh2023human}.
By requiring a definitive yes/no recommendation, we create a clear and interpretable signal for bias. This design enables precise quantification of how vocal profiles may skew audio LLMs' recommendations, offering a critical mechanism to detect and mitigate potential harms before such systems are deployed in real-world clinical practice.

\subsection{MedVoiceBias Dataset Construction}
To systematically evaluate the influence of audio characteristics, we constructed a dataset designed to benchmark bias assessment in audio LLMs. This dataset construction is detailed below.

\subsubsection{Speaker Demographics}  
We sourced utterances from the Common Voice repository~\cite{ardila2020common} (release \texttt{cv-corpus-22.0-2025-06-20}). Using the provided metadata, we focused on two age strata: (i) speakers self-reporting ages 20--29 (\textit{young}) and (ii) speakers self-reporting ages $\geq$60 (\textit{old}). 
Within each cohort, we retained only those speakers whose perceived age and gender are unambiguous to human raters. Manual validation was essential, as a systematic comparison of 200 randomly sampled profiles revealed discrepancies between self-reported and acoustically perceived demographics in 23\% of cases (age: 18\%, gender: 5\%). From this process, we selected 12 speakers, balanced across both age and gender (6M/6F; 6 young/6 old), with demographic classifications confirmed by consensus among three annotators.

\subsubsection{Emotional Variability}  
To examine whether paralinguistic cues influence downstream bias, we augmented the corpus with controlled emotional renderings. Specifically, we used the Expresso dataset~\cite{nguyen2023expresso}, selecting six affective conditions: \textit{happy}, \textit{laughing}, \textit{sad}, \textit{confused}, \textit{enunciated}, and \textit{whisper}. 
Expresso includes two male and two female speakers self-identified as young, and we incorporated all four speakers expressing each of the six emotions available in the dataset.

\subsubsection{Voice Cloning}  
In total, we constructed 36 unique voice profiles. For each selected speaker profile (demographic or emotional), we used Sesame-1B~\cite{sesame2024csm1b} to synthesize patient speech from DDXPlus~\cite{fansi2022ddxplus} clinical contexts, adapted to match the target voice characteristics. 
Because input length constraints degraded synthesis quality, we segmented the original patient profiles at the sentence level. For each sentence, we generated three candidate synthesis samples per voice profile and applied Whisper-v3~\cite{radford2023robust} to perform ASR on all outputs. The sample with the lowest word error rate (WER) was selected as the final audio input for downstream experiments, yielding an average WER of 6.4\% across all profiles. The statistics of our MedVoiceBias data is detailed in Table~\ref{tab:cohort_results}.

To further validate synthesis quality, we applied the MOSA-Net+~\cite{zezario2024study} automatic speech quality assessment model. The generated samples achieved average PESQ and intelligibility scores of 3.6/5.0 and 0.97, respectively, indicating consistently high-quality audio for subsequent evaluation.


\begin{table}[t]
\small
\centering
\caption{Statistics of our created MedVoiceBias data. $N$ represents the number of unique voice profile.}
\label{tab:cohort_results}
\vspace{1mm}
\begin{tabular}{llccc}
\toprule
\textbf{Category} & \textbf{Cohort} & \textbf{WER (\%)} & \textbf{Length (s)} & \textbf{N} \\
\midrule
\multirow{2}{*}{\textit{Age}} & Young & 6.1 & 34.0 & 10 \\
 & Old & 8.9 & 42.4  & 6 \\
\midrule
\multirow{2}{*}{\textit{Gender}} & Female & 6.6 & 35.7  & 10 \\
 & Male & 7.3 & 37.3 & 10 \\
\midrule
\multirow{6}{*}{\textit{Expression}} & Happy & 5.1 & 31.0 & 4 \\
 & Laughing & 5.3 & 33.7 & 4 \\
 & Sad & 5.6 & 33.7 & 4 \\
 & Confused & 5.6 & 36.5 & 4 \\
 & Enunciated & 6.3 & 38.9 & 4 \\
 & Whisper & 7.8 & 37.8 & 4 \\
\bottomrule
\end{tabular}
\vspace{-4.5mm}
\end{table}


\subsection{Bias Evaluation Metrics}

We evaluate bias by systematically comparing a model's clinical recommendations from \textbf{voice-based} inputs against those generated from a \textbf{text-only} control baseline. This approach allows us to isolate and quantify the influence of paralinguistic and demographic cues present in the synthesized audio.

Our primary metric is the \textit{surgery recommendation rate}, defined as the proportion of cases in which a model recommends a surgical intervention.
For each model and prompting strategy, we first establish this rate using the text-only baseline. We then calculate the recommendation rate for each distinct demographic cohort in our dataset (age, gender, and emotion), enabling a direct comparison of the model's behavior with and without audio cues.

All statistical comparisons utilize Fisher's exact test~\cite{upton1992fisher} to compare surgery recommendation rates between audio-based cohorts and the text-only baseline. We chose this test over Chi-squared to avoid issues with low expected cell counts in our demographic subgroups, ensuring precise statistical inference and exact p-values. A difference is considered statistically significant if the p-value is less than 0.05. To focus on clinically meaningful effects, we additionally require an absolute difference of 2\% or greater in the surgical recommendation rate to highlight a finding.

Prior to the main experiments, we performed two preliminary evaluations. First, we conducted a basic sanity check on the text-only patient profiles to ensure all models could correctly perform the fundamental task of providing a surgical recommendation.
Second, we fed audio from a Common Voice profile to each model to verify its ability to distinguish age and gender. This evaluation confirmed that the models could accurately recognize gender from voice, a key prerequisite for our bias analysis. This entire process was repeated for every model and prompting strategy to ensure a robust and comprehensive analysis.

\section{Experiments}
\label{sec:experiments}

To investigate implicit bias in audio LLMs for medical decision-making, we designed a two-phase experimental protocol. In the first phase, we validated each model's ability to (1) identify demographic characteristics from voice and (2) generate accurate surgical recommendations from text. In the second phase, we compared surgical recommendation rates across different input modalities (text vs. audio) and demographic groups.

\subsection{Setting}

We evaluated six state-of-the-art audio LLMs: DeSTA2.5-Audio 8B~\cite{lu2025desta2}, Qwen2.5-Omni 3B and 7B~\cite{Qwen2.5-Omni}, Gemini Flash (2.0, 2.5)~\cite{comanici2025gemini}, and GPT-4o-mini-audio~\cite{hurst2024gpt}. Before examining potential bias, we first confirmed that these models could reliably distinguish age, gender, and emotions from audio, as well as achieve performance above random chance when making surgical recommendations from the original text.  

Results in Table~\ref{tab:audio_age_gender_emotion} show varying levels of demographic detection across models. Gender prediction accuracy ranged from 96.1\% to 99.9\% for most models (with the exception of GPT-4o-mini at 0\%), while age prediction exhibited wider variance (32.6\% to 85.7\%). This demographic detection ability is a prerequisite for analyzing bias, as models are capable of perceiving demographic cues and then exhibit differential behavior.

For each model, we conducted experiments under two conditions: \textit{direct answer (DA)} and \textit{diagnose-then-decide chain-of-thought (CoT)}. In the DA setting, the model received the full patient profile (either audio or ASR transcripts) and was asked to directly output a binary decision regarding surgical necessity. 
In the CoT setting, the model was first prompted to infer the possible disease and then determine whether surgery is required.

\begin{table}[t]
\small
\centering
\caption{Audio model performance in text-mode surgery accuracy, voice-based age/gender/emotion identification accuracy (\%).}
\label{tab:audio_age_gender_emotion}
\vspace{1mm}
\begin{tabular}{l|c|ccc}
\toprule
\textbf{Model} & \textbf{Surgery} & \textbf{Age } & \textbf{Gender} & \textbf{Emotion} \\
\midrule
\textbf{gpt-4o-mini} & 76.2 & ~~0.0 & ~~0.0 & ~~0.0 \\
\textbf{gemini-2.0-flash} & 68.3 & 66.0 & 99.5 & ~~0.2 \\
\textbf{gemini-2.5-flash} & 55.5 & 57.4 & 99.9 & 17.0 \\
\textbf{Qwen2.5-Omni-3B} & 63.9 & 66.1 & 96.1 & 12.2 \\
\textbf{Qwen2.5-Omni-7B} & 60.3 & 66.1 & 97.5 & 16.9 \\
\textbf{DeSTA2.5} & 57.8 & 65.4 & 99.5 & 40.5 \\
\bottomrule
\end{tabular}
\end{table}


\begin{table}[t]
\small
\centering
\caption{Surgery recommendation rates without emotional expressions (\%).
\textbf{Bold} fonts indicate statistically significant differences (p \lt 0.05) compared to Text baseline.}
\label{tab:neutral_audio_comparison}
\vspace{1mm}
\begin{tabular}{l|cccc|cccc}
\toprule
& \multicolumn{4}{c|}{\textbf{Direct Answer (DA)}} & \multicolumn{4}{c}{\textbf{Chain-of-Thought (CoT)}} \\
\cmidrule(lr){2-5} \cmidrule(lr){6-9}
\textbf{Model} & \textbf{Text} & \textbf{Text+Profile} & \textbf{ASR} & \textbf{Audio} & \textbf{Text} & \textbf{Text+Profile} & \textbf{ASR} & \textbf{Audio} \\
\midrule
\textbf{gpt-4o-mini} & 26.5 & 26.5 & 19.4 & ~~\textbf{5.3} & 14.7 & 14.7 & 11.2 & 12.4 \\
\textbf{gemini-2.0-flash} & ~~0.0 & ~~0.0 & \textbf{14.1} & ~~0.6 & ~~7.6 & ~~7.6 & ~~6.5 & ~~6.5 \\
\textbf{gemini-2.5-flash} & 27.6 & 27.6 & 21.2 & 31.8 & ~~6.7 & ~~6.7 & 23.5 & 18.2 \\
\textbf{Qwen2.5-Omni-3B} & 97.6 & 97.6 & \textbf{14.8} & \textbf{75.3} & 31.8 & 31.8 & \textbf{15.4} & 35.9 \\
\textbf{Qwen2.5-Omni-7B} & 11.2 & 11.2 & ~~5.3 & \textbf{20.6} & 22.7 & 22.7 & 26.5 & 27.6 \\
\textbf{DeSTA2.5} & 53.9 & 53.9 & \textbf{26.5} & \textbf{88.8} & 26.8 & 26.8 & 28.3 & 28.5 \\
\bottomrule
\end{tabular}
\end{table}

\subsection{Modality-Induced Bias: Text vs. Audio}
\label{ssec:modality_bias}

Table~\ref{tab:neutral_audio_comparison} reveals substantial modality-dependent bias in surgical recommendations. Under the DA condition, 66.7\% (4 of 6) evaluated models exhibited statistically significant differences between text and audio inputs, with recommendation rate shifts ranging from -22.2\% for Qwen2.5-3B to +34.9\% for DeSTA2.5. 
A striking example is GPT-4o-mini, whose recommendation rate dropped from 26.5\% with text to just 5.3\% with audio, a relative reduction of 80\%, highlighting strong susceptibility to paralinguistic cues.

The ASR condition, which uses transcripts without paralinguistic features, showed intermediate levels of bias. For example, Qwen2.5-3B produced recommendation rates of 14.8\% (ASR) versus 97.6\% (text) and 75.3\% (audio), suggesting that both transcription errors and vocal characteristics contribute to disparities. Despite relatively strong ASR accuracy (average WER = 6.4\%), the ASR condition still produced significant deviations in 3 of 6 models under the DA setting, highlighting how even minor transcription errors can cascade into clinically meaningful differences in decision-making.

\subsection{Demographic-Aware Effects}
\label{ssec:demographic_bias}
We investigate how demographic cues influence the audio LLMs' clinical decisions to determine if these models are sensitive to profiles (e.g., more conservative for elderly patients).

\begin{table}[t]
\small
\centering
\caption{Age and gender effects in audio surgery recommendation rate without emotional expressions (\%).
\textbf{Bold} fonts indicate statistically significant differences (p < 0.05) between groups.}
\label{tab:bias_comparison}
\vspace{1mm}
\begin{tabular}{l|cccc|cccc}
\toprule
& \multicolumn{4}{c|}{\textbf{Direct Answer (DA)}} & \multicolumn{4}{c}{\textbf{Chain-of-Thought (CoT)}} \\
\cmidrule(lr){2-5} \cmidrule(lr){6-9}
\textbf{Model} & \textbf{Young} & \textbf{Old} & \textbf{Male} & \textbf{Female} & \textbf{Young} & \textbf{Old} & \textbf{Male} & \textbf{Female} \\
\midrule
\textbf{gpt-4o-mini} & 3.6 & 3.6 & \textbf{3.9} & \textbf{2.6} & \textbf{8.4} & \textbf{5.4} & 5.0 & 5.0 \\
\textbf{gemini-2.0-flash} & 0.7 & 0.6 & 0.6 & 0.5 & \textbf{6.0} & \textbf{3.7} & 3.7 & 3.5 \\
\textbf{gemini-2.5-flash} & \textbf{25.3} & \textbf{17.9} & 19.7 & 18.8 & \textbf{16.1} & \textbf{8.5} & 10.1 & 9.4 \\
\textbf{Qwen2.5-Omni-3B} & \textbf{85.3} & \textbf{73.5} & \textbf{76.7} & \textbf{73.2} & \textbf{23.7} & \textbf{28.2} & 30.0 & 28.1 \\
\textbf{Qwen2.5-Omni-7B} & \textbf{16.8} & \textbf{14.9} & 14.3 & 15.7 & \textbf{25.8} & \textbf{22.6} & 22.8 & 22.4 \\
\textbf{DeSTA2.5} & \textbf{87.6} & \textbf{90.1} & \textbf{93.5} & \textbf{83.7} & 22.6 & 20.9 & 20.9 & 18.9 \\
\bottomrule
\end{tabular}
\end{table}

\subsubsection{Age-Related Disparities}
Table~\ref{tab:bias_comparison} presents the surgical recommendation rates stratified by patient age. Contrary to our hypothesis that explicit reasoning would compel models to focus solely on the disease, CoT prompting revealed persistent and sometimes amplified age-related disparities.

In the DA condition, 4 of 6 models showed recommendations that significantly varied by age. Qwen2.5-3B showed the largest difference, recommending surgery for 85.3\% of young vs. 73.5\% of elderly patients.
This 11.8\% gap is a clinically meaningful finding that could lead to the systematic under-treatment of elderly patients.

Unexpectedly, CoT prompting increased the \textit{prevalence} of age-related differences, with five of six models showing significant variations. While the magnitude of these differences slightly decreased on average (mean absolute difference: DA = 4.9\%, CoT = 3.7\%), their increased consistency across models suggests that explicit reasoning may activate shared yet problematic clinical heuristics about age and surgical risk. Interestingly, models like DeSTA2.5 and Qwen2.5-3B reversed their recommendation patterns between the DA and CoT conditions, indicating that the reasoning pathways fundamentally alter how age cues influence a model's decisions. The findings imply that current audio LLMs are not yet equipped to effectively process and appropriately handle paralinguistic signals.

\subsubsection{Gender-Related Bias}
Gender bias patterns differed markedly from age-related disparities, as shown in Table 1. In the Direct Answer (DA) condition, only half of the models exhibited significant gender bias. The absolute differences in recommendation rates (ranging from 1.9\% to 8.0\%) were substantially smaller than those observed for age-related disparities.
DeSTA2.5 showed the largest gender gap, recommending surgery for 92.5\% of male vs. 84.5\% of female patients.

CoT prompting eliminated gender bias across all models, with no significant differences observed. This complete mitigation contrasts sharply with the age-related effects under CoT. This finding suggests that models may encode and process age and gender information through fundamentally different mechanisms, and that they are better equipped to handle gender cues than age cues.

\subsection{Emotional Expression Effect}
\label{ssec:emotion_bias} 

We evaluated whether emotional expression in speech affected surgery recommendations across six emotion categories.
Table \ref{tab:expresso_emotion_bias_da} shows the percentage of ``Yes'' recommendations for each emotion in the DA setting.
Most models exhibited relatively consistent behavior across emotions, with only two models showing significant differences: gemini-2.0 and DeSTA2.5. For gemini-2.0, the ``happy'' emotion (1.8\%) showed a notably higher recommendation rate compared to other emotions (0.3\% to 0.8\%).

However, the minimal variation in surgery recommendations across emotional expressions should be interpreted with caution. Given that most models demonstrated extremely low emotion detection accuracy (below 17\%), the observed consistency likely reflects their inability to perceive emotional cues rather than deliberate emotional robustness. 
Therefore, only results from models with demonstrated emotion detection capabilities, such as DeSTA 2.5, can provide meaningful insights into genuine emotion-based bias.

\begin{table}[t]
\small
\centering
\caption{Positive rate across all expressions in the DA setting (\%).
}
\label{tab:expresso_emotion_bias_da}
\vspace{1mm}
\setlength{\tabcolsep}{4pt} 
\begin{tabular}{l|rrrrrr|r}
\toprule
\textbf{Model} & \textbf{Conf} & \textbf{Enun} & \textbf{Hap} & \textbf{Lau} & \textbf{Sad} & \textbf{Whi} & Ref \\
\midrule
gpt-4o-mini & 3.8 & 4.6 & 4.2 & 4.8 & 3.6 & 3.8 & 26.5 \\
gemini-2.0 & 0.8 & 0.8 & 1.8 & 0.5 & 0.5 & 0.3 & 0.0\\
gemini-2.5 & 29.2 & 27.8 & 27.0 & 29.5 & 29.7 & 27.8 & 27.6\\
Qwen2.5-3B & 92.0 & 91.2 & 92.3 & 91.3 & 91.8 & 89.8  & 97.6\\
Qwen2.5-7B & 17.3 & 16.8 & 20.3 & 17.5 & 16.8 & 18.2  & 11.2 \\
\textbf{DeSTA2.5} & 90.3 & 87.4 & 84.7 & 87.8 & 92.5 & 87.9  & 53.9\\
\bottomrule
\end{tabular}
\vspace{-4.5mm}
\end{table}

\section{Conclusion}
\label{sec:conclusion}
This study provides the first systematic evaluation of voice-based bias in audio LLMs for medical decision-making. We created \textbf{MedVoiceBias}, which uses 170 clinical cases and 36 synthesized voice profiles to enable a controlled bias assessment.
Our findings reveal that audio LLMs exhibit significant instability, with surgery recommendation rates deviating a lot from text-only baselines. We demonstrate that these biases manifest across all demographic groups, with emotional expressions further amplifying the effect. While explicit reasoning from the models mitigates these biases, it does not eliminate them. This highlights a fundamental architectural challenge: the inability to reliably disentangle a patient's medical information from the paralinguistic features of their voice.
These results have critical implications for the deployment of speech-enabled AI in healthcare. Audio LLMs currently pose unacceptable risks of creating disparities based on how patients sound rather than on their medical needs. We conclude that bias-aware training and architectural innovations are imperative before clinical deployment to ensure that decisions are driven by medical evidence, not by a patient's voice.